# Generating Rescheduling Knowledge using Reinforcement Learning in a Cognitive Architecture

Jorge Palombarini[1,2], Juan Cruz Barsce[2], Ernesto Martinez[3]

[1] GISIQ(UTN) - Fac. Reg. Villa María, Av. Universidad 450, 5900, Villa María, Argentina.
`jpalombarini@frvm.utn.edu.ar`
[2] Dpto. Ing. en Sistemas de Información (UTN) - Av. Universidad 450, X5900 HLR, Villa María, Argentina.
`jbarsce@frvm.utn.edu.ar`
[2] INGAR(CONICET-UTN), Avellaneda 3657, S3002 GJC, Santa Fe, Argentina.
`ecmarti@santafe-conicet.gob.ar`

**Abstract.** In order to reach higher degrees of flexibility, adaptability and autonomy in manufacturing systems, it is essential to develop new rescheduling methodologies which resort to cognitive capabilities, similar to those found in human beings. Artificial cognition is important for designing planning and control systems that generate and represent knowledge about heuristics for repair-based scheduling. Rescheduling knowledge in the form of decision rules is used to deal with unforeseen events and disturbances reactively in real time, and take advantage of the ability to act interactively with the user to counteract the effects of disruptions. In this work, to achieve the aforementioned goals, a novel approach to generate rescheduling knowledge in the form of dynamic first-order logical rules is proposed. The proposed approach is based on the integration of reinforcement learning with artificial cognitive capabilities involving perception and reasoning/learning skills embedded in the Soar cognitive architecture. An industrial example is discussed showing that the approach enables the scheduling system to assess its operational range in an autonomic way, and to acquire experience through intensive simulation while performing repair tasks.

## 1 Introduction

Ensuring highly efficient production in an increasing dynamic and turbulent environment without sacrificing cost effectiveness, product quality and on-time delivery has become a key skill for surviving and prosper in the age of globalization. In this context, real-time rescheduling is a core capability because it enables a production system to react to unforeseen events like equipment failures, quality tests that demand reprocessing operations, delays in material inputs from previous operations, and the arrival of new orders that constantly demand repair operations by setting different schedule repair goals related to customer satisfaction and process efficiency, to guarantee due date compliance of orders-in-progress and negotiating delivery conditions [1].

As a result of the computational complexity of scheduling problems (NP-Hard) and the stochastic nature of industrial environments, schedules generated under the deterministic assumption are often suboptimal or even infeasible [2,3,4], so reactive scheduling is heavily dependent on the capability of generating and representing knowledge about strategies for repair-based scheduling which can be used in real-time for producing satisfactory schedules rather than optimal ones in a reasonable computational time [2]. In this way, many research efforts have identified the necessity of developing novel rescheduling methodologies based on the integration of human-like cognitive capabilities [3,4,5] along with learning/reasoning skills and general intelligence to generate rescheduling knowledge which is required to counteract unexpected situations and unforeseen events in real-time, in order to achieve higher degrees of flexibility, adaptability and autonomy in production systems [5,6].

In this work, a novel real-time rescheduling approach is presented, which resorts to a general cognitive architecture capabilities and integrates symbolic representations of schedule states with (abstract) repairing operators. To learn a near-optimal policy for rescheduling, using simulated schedule state transitions, an interactive repair-based strategy bearing in mind different goals and scenarios is proposed. In that sense, domain-specific knowledge for reactive scheduling is developed and integrated with learning mechanisms in the Soar cognitive architecture such as chunking and reinforcement learning via long term memories [7]. Finally, an industrial example is discussed showing that the approach enables the rescheduling system to cope with disturbances and unplanned events in an autonomic way, and to further acquire rescheduling knowledge in the form of production rules through intensive simulation.

## 2 Real-time Rescheduling in Soar Cognitive Architecture

Knowledge about heuristics for repair-based scheduling to deal with unforeseen events and disturbances can be generated and represented by resorting to the integration of a schedule state simulator with Soar cognitive architecture [7]. In the simulation environment, an instance of the schedule is interactively modified by the system using a sequence of actions called *repair operators* suggested by Soar, until either a repair goal is achieved or the impossibility of repairing the schedule is accepted.

In this way, the cognitive architecture is connected with the rescheduling component via a .NET wrapper [5]. The symbolic first order information about actual schedule state comes in from the Soar perception module to its working memory, which cues the retrieval of rescheduling knowledge from Soar's symbolic long-term memories. Procedural long-term memory stores knowledge represented as *if-then* production rules, which match conditions against the contents of the working memory to control the architecture execution cycle. To control the rescheduling behavior, they generate preferences, which are used by the decision procedure to select a repair operator. Operators are key components in the proposed approach because they can be applied causing persistent changes to the working memory, which has reserved areas that are monitored by other memories and processes, whereby changes in the working memory can either initiate retrievals from both Semantic and Episodic memories, or initiate schedule repair actions through the abstract actuator in the environment. The Semantic memory stores general first order structures that can be employed to solve new situations; i.e. if in the schedule state *s1* the relations `precedes(task1,task2)` and `precedes(task2,task3)` exist, the Semantic memory can add the abstract relation `precedes(A,B), precedes(B,C)` to generalize such knowledge. On the other hand, the Episodic memory stores streams of experience in the form of *state-operator...state-operator* chains to predict the environmental dynamics in similar situations, or to envision the schedule state outside the immediate perception. Moreover, this approach uses two specific learning mechanisms associated with Soar's pro-cedural memory: reinforcement learning and chunking [8]. Finally, repair operators suggested by the decision procedure affect the schedule state and are provided to the real-time rescheduling component via an integrated OutputLink structure.

### 2.1 Schedule States and Repair Operators representation in Soar

Soar's working memory holds the current schedule state, and it is organized as a connected graph structure via a semantic net rooted in a symbol that represents the state, as it is shown in Fig. 1. The non-terminal nodes of the graph are called identifiers whereas the arcs, the attributes, and the concrete values are the terminal nodes. The arcs that share the same node are called objects, which consist of all the properties and relations of an identifier. A state is an object, and all other objects are substructures of the state, either directly or indirectly. Fig. 1 provides an example of a state named `<s>` in Soar's working memory (some details have been omitted for the sake of clarity).

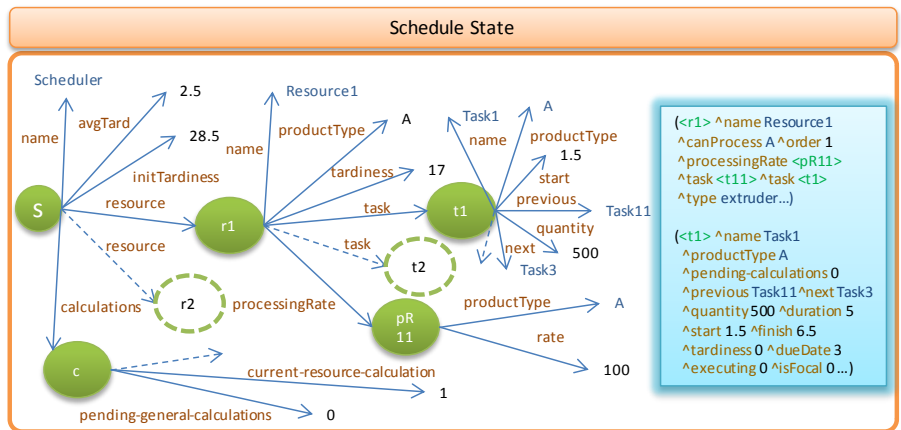

**Fig. 1.** Partial representation of a schedule state as a semantic net (left), and partial symbolic specification of identifiers `<r1>` and `<t1>` in Soar

In Fig.1, the actual state is called `<s>`, and it has attributes like avgTard with the value 2.5 h., initTardiness with the value 28.5 h. and a totalWIP of 46.83, among others. Also, there exists a resource attribute, whose value is another identifier, so it links the state `<s>` with other objects called `<r1>`, `<r2>`. `<r1>` is a resource of the type extruder, which can process products of type A, that has an accumulated tardiness in the resource of 17 h., and has two tasks assigned to it: `<t1>` and `<t2>`. They are in turn objects, in such a way that, for instance, `<t1>` have proper attributes which describe it like Product Type, Previous Task, Next Task, Quantity, Duration, Start, Finish and Due Date, among others. Also, resources have an important attribute called Processing Rate that determines the total quantity of a type of product which a given resource can process in a unit of time. Finally, a state attribute called calculations stores data that are relevant to calculate task tardiness, resource tardiness, schedule total tardiness, and derived values. Such symbolic representation provides a higher level of flexibility, so the number of objects and attributes can change between states, which make their *a priori* definition unnecessary (See right part of Fig. 1).

In a rescheduling task, Soar is in a specific schedule state at a given time (represented in working memory) while attempting to select an operator so that will move it to a new state [8]. Such process continues recursively towards a goal state (i.e. the schedule with a total tardiness that is minor than the initial tardiness) through a number of cycles; where each cycle is applied in phases as follows: the **Input**, where new sensory data comes into the working memory; the **Elaboration**, where production rules fire (and retract) to interpret new data and generate derived facts; the **Proposal-Evaluation**, in which the architecture propose and evaluate repair operators for the current schedule state using preferences; and the **Application**, where the selected operator application rules fire. Finally, output commands are sent to the real-time rescheduling component until the halt action is issued from the Soar program.

## 3 Repair Operators Evaluation using Reinforcement Learning

Repair operators have been designed to change/swap the execution precedence of a task with other tasks in the schedule (which may be assigned to other resource), or to change/swap to an alternative resource (disperser, formulation unit, filling train) assignation so as to reach a goal state [5]. Each operator takes two arguments: focal task/resource, and auxiliary task/resource. The Focal task/resource is taken as the reparation anchorage point whereas an auxiliary task/resource specifies the reparation action and evaluates its effectiveness. For example, if the proposed operator is `Task-Down-right-swap-precedence`, its effect is interchanging the execution precedence between the Focal Task with a task programmed to start later in an alternative resource whose identifier is greater than the assigned to execute the Focal Task.

The conditions that must be met by the schedule state for the operator to be applied are hard-coded in the left hand side of the operator proposal rule (pre-conditions). In turn, the right hand side of the operator proposal rule defines how the schedule state will change as a consequence of the repair operator application (post-conditions). Once the repair operator and its arguments have been obtained, the rest of the variable values can be totally defined as they are related with each other and variables in the body rule act as universal quantifiers, so the repair operator definition and application can match completely different schedule situations and production types only with the relational restrictions established in the left hand side of the rules.

In order to carry out the operator evaluation, control knowledge, which employs *numeric preferences,* is used to select one of the proposed operators to be applied [9]. *Numeric preferences* specify the expected value of applying an operator in a particular state, so it is an estimation of the long term total reward to be obtained from its application in a particular state and is taken as a reference to compare several potential operators. Soar holds a structure in the working memory called `^reward.value` which stores the obtained reward after the application of an operator. The expected value is represented in Soar using operator evaluation rules which analyze and create *numeric preference*s (*RL rules)*. The latter match the schedule state and the proposed operators creating *numeric preferences* (See Example 1). The preference values are combined to obtain a unique value for evaluation purposes.

```
Example 1. sp {Scheduler*rl*rules*156 (state <s> ^name Scheduler
^initTotRatioScale   1   ^avgMaxRatioScale   3   ^tardWIPRatioScale   2
^relativeTardFocalScale  2  ^relativeTardAuxiliarScale  1^avgTard <avg-
Tard>> 3.56 ^cantTask <cantTask>< 16 ^initTardiness <initTardiness>>
23.5 ^maxTard <maxTard>> 19.2 ^resource <r1> ^resource <r2> ^totTard
<totTard>>  35.7  ^totalWIP  <totalWIP><  103.4  ^cantResource  <cantRe-
source>> 2 ^operator <o> +)(<o>    ^name down-right-jump ^focalTask
```

```
<nametFocal>^auxTask <nametAux>) (<r2> ^tardiness <res2Tard> ^task <tFo-
cal>) (<r1> ^tardiness <res1Tard>><res2Tard> ^task <tAux>) (<tAux> ^name
<nametAux>)(<tFocal> ^name <nametFocal>)
   --> (<s> ^operator <o> = 0.184)}
```

Reinforcement learning involves modifying numeric preference values iteratively, stored in *RL rules*. In this work, a SARSA(λ) algorithm is used to update preference values after each operator application based on *eligibility traces* [10] to speed-up convergence. After the changes in the schedule state have been performed by the repair operator application rule, Soar reinforcement learning rules are fired, so as the architecture can learn in the form of numeric preferences from the results of the particular repair operator application. Such process is performed using Eq. (1). The reward function *r* is defined as the amount of tardiness decreased or increased.

$$Q(s,ro) = Q(s,ro) + \alpha[r + \gamma Q(s',ro') - Q(s,ro)]e(s,ro) \quad (1)$$

where $Q(s,ro)$ is the value of applying the repair operator *ro* in schedule state *s*, $\alpha$ and $\gamma$ are the learning rate and the discount factor respectively, and *r* is the reward value whereas $e(s,ro)$ is the eligibility trace value of *ro* in *s*.

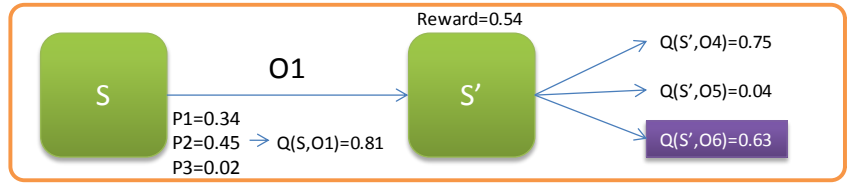

**Fig. 2.** Evaluation Rules: value updating using Reinforcement Learning

Fig. 2 shows a situation in which the operator O1 is selected in the state S based on *RL rules* P1, P2 and P3 whose combined values resulting in an expected value of 0.81. Once the operator has been applied, a reward of 0.54 is received, and three new operators are proposed: O4, O5 and O6. If O6 is selected, a preference update must be calculated (with parameters set as γ = 0.9, α = 0.1, and *e*=1) as it is shown in Eq. (2):

$$0.1[0.54 + 0.9 * 0.63 - 0.81] = 0.1[0.54 + 0.567 - 0.81] = 0.0297 \quad (2)$$

The corresponding value is divided between the rules which have contributed to its calculation, in this case P1, P2 and P3, each one of them adjusted in 0.01 (0.0297/3). Then, the automatic *chunking* process extends the learning with new production rules that can be executed in real-time starting from an initial schedule state [11].

## 4   Industrial Case Study

To illustrate our approach for automatic generation of first-order logic repair strategies, an industrial case study has been considered. In this manufacturing system, there exists a net of paint manufacturing plants, based on a five stage structure detailed in [12], which is comprised of more than 80 equipments in total, among formulation units, laboratory and filling-out trains. The processing stages consist of formulation and mixing, laboratory quality checking, container filling and palletization (See Fig. 3). Also, there are 100+ products differing mostly in product types (alkyds and latex or water-based), product families (where the container size is the only distinctive feature), packaging type, product size and batch size. The formulation and mixing steps were integrated into the processing activity, and it was considered that filling machines are connected to their respective packaging machines, palletizer and all necessary connecting conveyors which are made up of alternative fill-out trains. Set-up times for the cleaning and the reconfiguration are all included in the fill-out processing times. The equipment items used are virtual formulation units of different tank sizes with 11 units used for latex products, and 40 for alkyds. Each formulation unit consists of a disperser and a tank. After processing a batch, the corresponding disperser is disconnected from the tank so that a different formulation unit can be set up using a different tank.

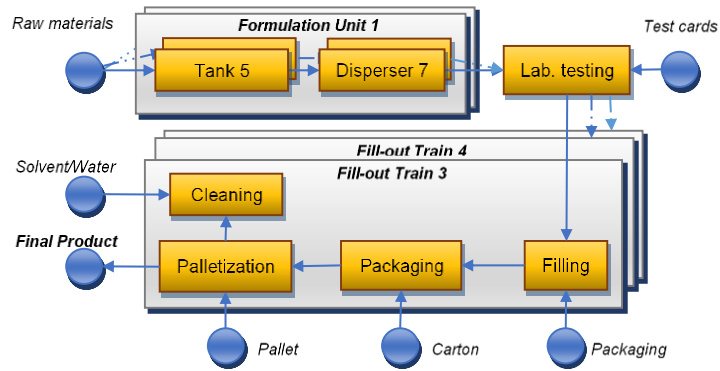

**Fig 3.** Paint production process

For the latex product line, there are 4 dispersers available, whereas for alkyds, there are 13. The tanks that share the same disperser can hold a product batch as a work-in-process, but cannot be simultaneously used for (re)formulating another product batch. Each product batch is kept in its corresponding tank until the packaging operation is completed via the chosen fill-out train.

Three applications have been used to implement and test this industrial case study: i) Visual Soar v4.6.1 to design and implement the definition of schedule state and operator structure, ii) Soar Debugger 9.3.2 to test and run mentioned rules and to perform the train of the reschedule agent [13], and iii) Visual Studio 2010 to develop the real time scheduling component, which allowed to validate the results and to read/write on the Soar output/input link, respectively.

The disruptive event that has been considered is the arrival of a new order and, for learning rescheduling rules, the SARSA($\lambda$) algorithm was used with an $\varepsilon$-greedy policy, eligibility traces, $\gamma = 0.9$, $\varepsilon = 0.1$, $\lambda = 0.1$ and $\alpha = 0.1$, which are standard values for such parameters [10]. As it can be seen in Fig. 4, 1000 simulated episodes were executed in the training phase of the agent, and three repair goals have been considered: Tardiness Improvement, Stability and Balancing, which differ in the definition of goal states, alternatively prioritizing efficiency, stability or a mix between both [14]. Each episode is comprised of a set of state transitions, starting in the initial schedule state, and finalizing in a goal schedule state which is previously unknown (e.g. a state with a Total Tardiness less than 1 working day). As a result of the training process, it can be observed that the most costly goal for the event is Tardiness Improvement (blue curve), which requires on average more than 800 repair operator applications in the first 100 episodes. Later on, the process tends to stabilize on an average of 80 operators applications, with peaks between the episodes #350 and #400.

The Tardiness Improvement goal converges to a lower average of the operator applications required to achieve a repair goal compared to Balancing (green curve), which achieves convergence with an average of about 120 operators without noticeable peaks. Finally, the Stability goal (red curve) requires, on average, 50 steps to repair a schedule. Moreover, as it is shown in the right section of Fig. 3, Soar has generated approximately 200.000 *RL rules*. The greatest amount corresponds to Balancing, followed by Tardiness Improvement and then by Stability. Furthermore, for the first two goals in the 100 initial episodes, almost one rule per operator is generated until the automatic creation of structures in the Semantic Memory via chunking takes effect, retracting almost 28.000 rules. Also, the Stability goal seems to be the less costly computationally-wise in terms of knowledge generation, storage and recovery.

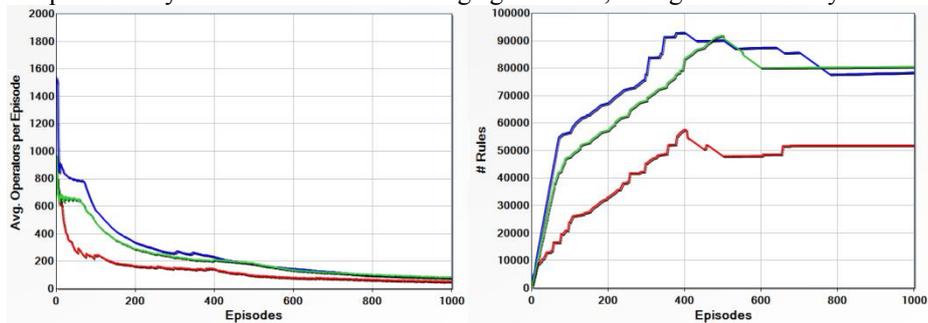

**Fig. 4.** Training process results for Arrival of a New Order event: Accumulated Avg. Operators per Episode (left), and Total Number of Rules per Episode, generated by Soar (right)

In order to apply the learned repair policy, an initial schedule has been generated using a FIFO dispatch rule, from which an operator repair sequence suggested by Soar *RL rules* has been applied.

**Table 1.** Results after the application of the learned RL rules for Tardiness Improvement goal.

|  | Initial Schedule | Schedule post insertion | Repaired Schedule |
|---|---|---|---|
| Total WIP (h) |  | 1490 |  |
| Tardiness (h) | 171,90 | 344,65 | 147,92 |
| Exceeded Tasks (#) | 21 | 59 | 20 |
| Avg. Tard.(h/t) | 1,63 | 3,28 | 1,40 |

Results obtained are shown in Table 1, which provides the Total WIP (Work In Process) and the Total Tardiness (h), which is the total number of tasks which exceed the original Due Date, and the Average Tardiness (h/t) for both, the initial and the repaired schedule, as well as for the schedule obtained after the insertion of the arriving batch for Tardiness Improvement goal. Such event produces an increment of nearly 100% in the Total and the Average Tardiness, tripling the number of tasks whose due dates have been exceeded. After applying the repair policy, the new batch has been inserted reducing the Initial Tardiness in 13%, whilst tasks whose due dates are exceeded have been reduced in 1 respect to the original schedule. It is important to note that in real industrial environments, once the training process has been carried out in an off-line way through simulated state transitions, the rescheduling knowledge stored in the form of logical rules can be applied on-line to repair new schedules without extra computation, so there is no need to re-train the rescheduling agent every time a new schedule needs to be repaired.

## 5 Concluding Remarks

A novel strategy for simulation-based learning of a rule-based policy has been presented, which deals with automated repair in real time schedules using reinforcement learning in the Soar cognitive architecture. The policy generates a sequence of local repair operators to achieve rescheduling goals in order to handle abnormal and unplanned events such as inserting an arriving order with minimum tardiness based on a symbolic first order representation of schedule states using repair operators. The proposal efficiently use and represent large bodies of symbolic knowledge, as it dynamically combines available knowledge for decision-making in the form of production rules with learning mechanisms, and it can compile the rescheduling problem solving into production rules, so that the schedule repair process is replaced by rule-driven decision making which can be used reactively in real-time in a straightforward way. Finally, relying on an appropriate and well-designed set of template rules, the approach enables the automatic generation through reinforcement learning and chunking of rescheduling heuristics that can be naturally understood by an end-user.